\documentclass[conference]{IEEEtran}
\IEEEoverridecommandlockouts
\usepackage{cite}
\usepackage{amsmath,amssymb,amsfonts}
\usepackage{algorithm} 
\usepackage{algorithmic}   
\usepackage{subcaption} 
\usepackage{graphicx}
\usepackage{multirow}
\usepackage{diagbox}
\usepackage{textcomp}
\usepackage{xcolor}
\usepackage{CJKutf8}

\def\BibTeX{{\rm B\kern-.05em{\sc i\kern-.025em b}\kern-.08em
    T\kern-.1667em\lower.7ex\hbox{E}\kern-.125emX}}
\begin{document}

\title{Generation of Synthetic Electronic Medical Record Text\\
}

\renewcommand{\thefootnote}{\fnsymbol{footnote}}
\author{\IEEEauthorblockN{Jiaqi Guan}
\IEEEauthorblockA{\textit{Department of Automation} \\
\textit{Tsinghua University}\\
Beijing, China \\
guanjq14@tsinghua.org.cn}
\and
\IEEEauthorblockN{Runzhe Li}
\IEEEauthorblockA{\textit{Department of Mathematical Sciences} \\
\textit{Tsinghua University}\\
Beijing, China \\
lrz14@tsinghua.org.cn}
\and
\IEEEauthorblockN{Sheng Yu}
\IEEEauthorblockA{\textit{Center for Statistical Science} \\
\textit{Department of Industrial Engineering} \\
\textit{Institute for Data Science} \\
\textit{Tsinghua University}\\
Beijing, China \\
syu@tsinghua.edu.cn}
\and
\IEEEauthorblockN{Xuegong Zhang\footnotemark*\textit{, Member, IEEE}}
\IEEEauthorblockA{\textit{Department of Automation} \\
\textit{Tsinghua University}\\
Beijing, China \\
zhangxg@tsinghua.edu.cn}
}

\maketitle
\renewcommand{\thefootnote}{\fnsymbol{footnote}}
\footnotetext[1]{Corresponding author}

\begin{abstract}
Machine learning (ML) and Natural Language Processing (NLP) have achieved remarkable success in many fields and have brought new opportunities and high expectation in the analyses of medical data. The most common type of medical data is the massive free-text electronic medical records (EMR). It is widely regarded that mining such massive data can bring up important information for improving medical practices as well as for possible new discoveries on complex diseases. However, the free EMR texts are lacking consistent standards, rich of private information, and limited in availability. Also, as they are accumulated from everyday practices, it is often hard to have a balanced number of samples for the types of diseases under study. These problems hinder the development of ML and NLP methods for EMR data analysis. To tackle these problems, we developed a model to generate synthetic text of EMRs called Medical Text Generative Adversarial Network or mtGAN. It is based on the GAN framework and is trained by the REINFORCE algorithm. It takes disease features as inputs and generates synthetic texts as EMRs for the corresponding diseases. We evaluate the model from micro-level, macro-level and application-level on a Chinese EMR text dataset. The results show that the method has a good capacity to fit real data and can generate realistic and diverse EMR samples. This provides a novel way to avoid potential leakage of patient privacy while still supply sufficient well-controlled cohort data for developing downstream ML and NLP methods. It can also be used as a data augmentation method to assist studies based on real EMR data. 
\end{abstract}

\begin{IEEEkeywords}
synthetic EMR text, conditional model, generative adversarial network, reinforcement learning
\end{IEEEkeywords}

\section{Introduction}
The widespread adoption of Electronic Medical Records (EMR) has brought new opportunities in the biomedical domain, of which clinical narratives are significant components. However, due to relevant laws and regulations for the protection of patient privacy, EMR data are generally inaccessible to the majority of the ML community. In addition, when studying a disease, the positive and negative EMR samples are usually highly imbalanced, which makes it difficult to train ML algorithms with real medical records. As a result, we aim to develop a model to generate synthetic textual EMR datasets. 

A general way to alleviate the privacy risks is via de-identification, which is the process of reducing the information associated with an individual's identity. The anonymization is typically done by applying generalization and suppression operations to modify the patients' attributes \cite{b1}. However, these approaches cannot fully avoid privacy disclosure, since the anonymous patients can be re-identified using specific information\cite{b2}. Generating synthetic text of medical records is a way to completely avoid possible re-identifications. Text generation is one of the most fundamental problems in natural language processing. During the recent decade, deep neural networks have achieved remarkable success in several tasks and researchers are paying more attention to the text generation via deep learning models. A promising approach to text generation is training a recurrent neural network (RNN) by maximum likelihood estimation (MLE) \cite{b3}. However, it suffers from the well-known \emph{exposure bias} \cite{b4} problem. The success of Generative Adversarial Network (GAN) \cite{b5} has inspired researchers to investigate adversarial training over textual data, while another new problem has arisen when using GAN to generate discrete data: the gradient cannot be back-propagated from the discriminator to the generator. Some related work (such as SeqGAN \cite{b6}) utilize the REINFORCE algorithm \cite{b7}, which is a classical policy gradient algorithm in reinforcement learning, to optimize the original GAN objective. 

Our proposed model Medical Text Generative Adversarial Network (mtGAN) is a GAN-based framework and we adopt the REINFORCE algorithm to train the model. To satisfy different demands of research, our mtGAN is a conditional model with designated disease features as inputs, and can generate corresponding EMR text data. We test three discriminative models (fastText, CNN, BiRNN) and different methods to rescale rewards to achieve the best performance. We design micro-, macro- and application-level experiments to demonstrate the effectiveness of our model. In the micro-level experiment, our model has the strong ability to fit the real data and generate diverse examples at the same time. In the macro-level experiment, our model has the best adversarial success in the adversarial evaluation. In the application-level experiment, we design a disease classification experiment and the results suggest that the synthetic EMR texts generated by our model are capable of producing comparable properties to real data. Therefore our synthetic datasets can also be used as a data augmentation approach for machine learning tasks.

\section{Related Work}
Text generation has been one of the most challenging problems in natural language processing. RNN and its variants Long Short-Term Memory (LSTM) \cite{b8} and Gated Recurrent Unit (GRU) \cite{b9} have achieved impressive performance in several complex tasks, such as machine translation and dialogue generation\cite{b11}. The RNN language models are commonly used for sequence generation, and they are trained by MLE in an approach called \textit{teacher forcing} \cite{b12}. This mainstream method for auto-regressive models predicts the next token given the previous ground-truth tokens, which leads to the exposure bias problem.

 GAN proposed by Goodfellow provides an alternative framework to generate synthetic data. The GAN model consists of two neural networks: a generator G trying to generate synthetic data, and a discriminator D trying to distinguish the real data from the synthetic. The training procedure is a two-player zero-sum game between G and D. GANs have enjoyed great success in image generation, but they are not as much widely applied in natural language processing tasks. One reason is that the gradient from the discriminator cannot be back-propagated to the generator due to the discrete outputs. To address this problem, Yu proposed seqGAN \cite{b6}, where the generator is updated through the policy gradient using Reinforcement Learning (RL), and the reward is calculated by the discriminator on a complete sequence via Monte Carlo search. However, no one has ever used this kind of approach in the case of generating synthetic EMR text.

Recent studies attempt to generate synthetic electronic medical records via deep generative models. For example, Choi proposed a new model medGAN to generate realistic EMRs with high-dimensional binary and count variables \cite{b13}. Hyland proposed a Recurrent (Conditional) GAN to generate real-valued time series in medical application \cite{b14}. Yahi utilized a GAN framework to produce continuous time series data in EMRs, which can predict the effects of drug exposure \cite{b15}. However, most of the research on synthetic medical records are based on highly structured data formats, including numerical and categorical variables, while the majority of clinical documents are saved in unstructured textual formats. Under such circumstances, we propose a model called \textit{mtGAN} for the generation of synthetic EMR text. The primary contributions of our paper are listed as follows:
\begin{itemize}
\item We propose mtGAN to generate synthetic EMR text. \item We can control the generation process with assigned specific disease features to satisfy various research demands.
\item We design micro-level, macro-level and application-level evaluation methods to assess the model performance, and our model outperforms other baseline models.
\item The application-level experiment results demonstrate that our synthetic data can achieve similar performance in machine learning tasks compared with the real data, which can be used as a data augmentation approach then. 
\end{itemize}

\section{Preliminaries}
\subsection{Synthetic EMR Text Generation Problem}
The synthetic EMR text generation problem is fundamentally a discrete sequence generation problem. Specifically, suppose we have a set of real-world EMR data $S^+ = \{X^i\}_{i=1}^N$, where each data consists of a sequence of words $X = \{x_1, x_2, \dots, x_T\}$ and each word comes from a vocabulary of candidate tokens. Our goal is to produce a set of EMR data which have similar characteristics to real-world EMR data by learning the underlying distribution of the real data $p_d$, so that it can be used in more cases to substitute privacy-sensitive and limited real EMR data. 

\subsection{Recurrent Neural Networks}
In recent years, RNN and its improved variants, such as LSTM and GRU, have shown remarkable results in the text generation task. At each time step, RNN will encode previous inputs to a hidden vector $h_t$ and use it to conduct the inference of the next token. This procedure can be formulated as:
\begin{equation}
	\left\{
		\begin{aligned}
		    h_t &= f(h_{t-1}, x_t) \\
            o_t &= g(h_t) \\
		\end{aligned}
	\right.
\end{equation}

\subsection{Generating with Maximum Likelihood Estimation}
In the text generation task, the general training method is Maximum Likelihood Estimation (MLE), which converts the text generation problem to a sequential multi-label classification problem. For a RNN generator $G_\theta$, the MLE objective is to minimize the multi-label cross entropy, which can be formulated as \eqref{eq:MLE}. For the simplicity of notations, we will also denote the probability of generating tokens as $G_\theta(\cdot|\cdot)$. 
\begin{equation}
    \label{eq:MLE}
	J_G(\theta) = \mathbb{E}_{X\sim p_{d}}[-\sum_{t=1}^{T}\log G_\theta(x_t|X_{1:t-1})]
\end{equation}

\subsection{Generating with Adversarial Reinforcement Learning}
Although MLE has better convergence performance and training robustness than other algorithms, it suffers from the \emph{exposure bias} problem, which makes MLE less useful in generating long texts. To tackle this problem, recent works focus on generating texts under the GAN setting. In the standard GAN setting, there is a generator $G$ that plays a minimax game against a discriminator $D$. The generator $G$ transforms a noise $z$ sampled from a noise distribution $p_z$ to a data sample $G(z)$, and tries to match the generated distribution $p_g$ to the real data distribution $p_d$. The discriminator $D$ is a binary classifier, which takes real data as positive samples while synthetic data as negative samples, and tries to distinguish them. The two-player minimax game for continuous data can be formulated as follows, where $G(\cdot)$ denotes the generated sample, $D(\cdot)$ denotes the probability given by $D$ that the sample comes from the real distribution. 
\begin{equation}
	\label{eq:gan_goal}
	\begin{aligned}
		\min_G\max_D V(D,G) =
		&\mathbb{E}_{\mathbf{x}\sim p_{d}(\mathbf{x})}[\log D(\mathbf{x})] + \\
		&\mathbb{E}_{\mathbf{z}\sim p_{z}(\mathbf{z})}[\log (1-D(G(\mathbf{z}))]
	\end{aligned}
\end{equation}

It can be seen that the standard GAN framework requires the generated data is differentiable so that the gradient can back-propagate from D to G to update parameters of the model. This constraint makes it not trivial to apply the standard GAN framework to discrete data, such as EMR text data. One way to solve this problem is to apply a typical reinforcement learning (RL) algorithm to the generator. Text generation with RNNs can be viewed as a sequential decision process. Hence, we can model the generator as a policy of picking the next token. The key elements in RL are as follows: 
\begin{itemize}
    \item State: The generated tokens so far $x_{1:t-1}$
    \item Action: The next token to be generated $x_t$
	\item Reward: The GAN discriminator's output, i.e. the likelihood that the synthetic sentence can fool the discriminator, which indicates how good the entire sentence is.
\end{itemize}

The reward defined by the aforementioned way will be used for all actions (the generated tokens). However, in some cases, the discriminator might assign a low reward due to bad parts of the generated sentence, which is not fair for the good generated parts. Thus, rewards for intermediate generation steps are necessary. One simple strategy to compute intermediate rewards is using Monte Carlo (MC) search \cite{mc-search}. In MC search, given a partially generated sentence $X_{1:t-1}$, the model keeps sampling tokens from the current distribution until the sentence ends, and repeats this sampling procedure for $K$ times. These $K$ samples are fed to the discriminator, and the average score is used as the reward for $x_t$. 

With well-defined RL elements and intermediate rewards, we can optimize the model with the REINFORCE algorithm. Given a generator $G_\theta$ which tries to maximize the rewards it receives from the discriminator, and a discriminator $D_\phi$ which tries to distinguish real text from synthetic text, the objective can be formulated as \eqref{eq:g_obj} and \eqref{eq:d_obj}. For generating discrete data,  $G(x_t|X_{1:t-1})$ denotes the probability of generating token $x_t$ given previous generated tokens $X_{1:t-1}$.
\begin{small}
\begin{align}
	\label{eq:g_obj} \max J_G(\theta) \!&=\! \mathbb{E}_{X\sim G_\theta}[\sum_{t=1}^{T}\! \log G_\theta(x_t|X_{1:t\!-\!1}) \!\cdot\! R_{D_\phi}^{G_\theta}(X_{1:t\!-\!1}, x_t)] \\
	\label{eq:d_obj} \max J_D(\phi) \!&=\! \mathbb{E}_{X\sim p_{d}}[\log D_\phi(X)]\!+\!\mathbb{E}_{X\sim G_\theta}[\log (1\!-\! D_\phi(X))]
\end{align}
\end{small}
where, 
\begin{small}
\begin{equation}
	\label{eq:qvalue}
	R_{D_\phi}^{G_\theta}(s\!=\! X_{1:t\!-\!1},a\!=\! x_t)=
	\frac{1}{K} \sum_{k=1}^K D_\phi(X_{1:T}^k), X_{1:T} \in \text{MC}(X_{1:t}) 
\end{equation}
\end{small}

Our synthetic EMR text generation model will base on the GAN framework and use REINFORCE as our optimization algorithm.

\section{Method}
\subsection{Medical Text Generative Adversarial Network} 
\begin{figure*}[tb]
    \centering
    \subcaptionbox{\label{fig:mtGAN} mtGAN}
		{\includegraphics[width=8cm, height=3.8cm]{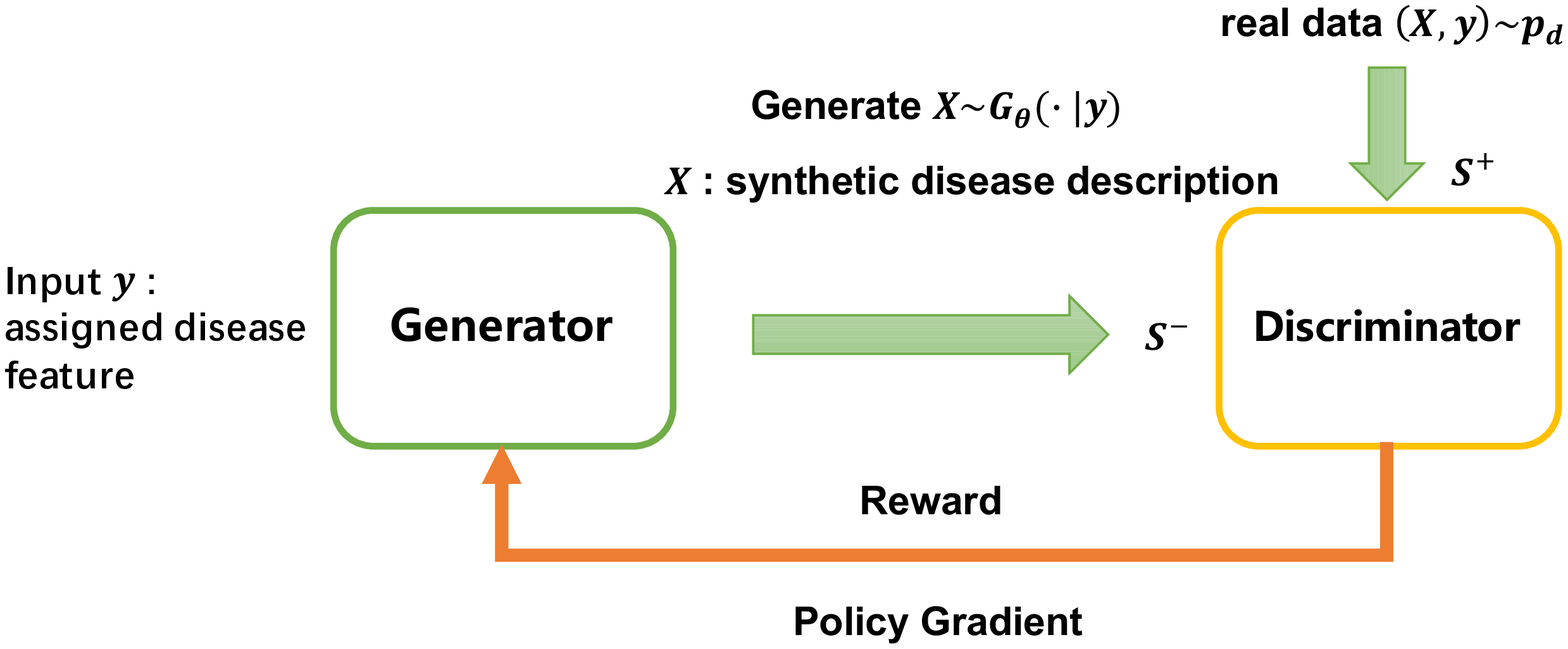}}
    \hspace{4em}
    \subcaptionbox{\label{fig:generator} Generator}
    	{\includegraphics[width=8cm, height=4.5cm]{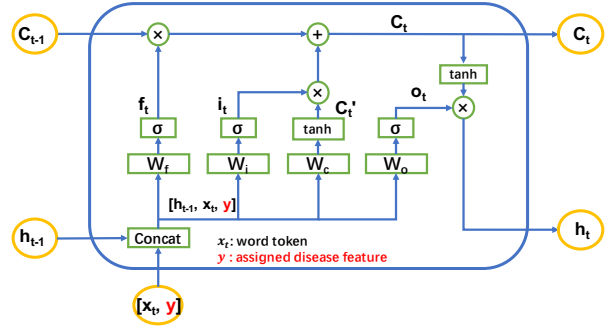}}
    \caption{Illustration of mtGAN and the conditional generator model}
\end{figure*}

To apply GAN to generate synthetic EMR text data, we propose a conditional GAN framework named Medical Text Generative Adversarial Network (mtGAN). In the medical domain, EMR text usually consists of disease descriptions and diagnostic results. The underlying mapping from disease descriptions to diagnostic results is what researchers essentially focus on. 
Our proposed mtGAN is a conditional GAN \cite{condgan} that takes designated disease features as inputs, and generates corresponding EMR text. For example, the disease features can be pneumonia or lung  cancer, which indicates the severity of the disease. When we train the mtGAN model, these features can be extracted from complete real EMRs. The overall mtGAN model is shown in Fig. \ref{fig:mtGAN}. We input disease features to guide the generator to produce corresponding synthetic disease descriptions, which are fed to the discriminator with real EMR text together. The classification results of D are the reward signals to guide the training of G.  A sketch of the training of mtGAN is shown in Algorithm \ref{algo:main}. It is worth noting that the introduction of conditional constraint is the main difference compared to seqGAN. More details will be described in the following subsections.

Given a generator $G_\theta$  and a discriminator $D_\phi$, we introduce an additional conditional constraint $y$ based on the original GAN framework, and the objectives of the generator and the discriminator can be rewritten from \eqref{eq:g_obj} \eqref{eq:d_obj} \eqref{eq:qvalue} to \eqref{eq:condg_obj} \eqref{eq:condd_obj} \eqref{eq:cond_qvalue}:
\begin{small}	
\begin{align}
	\label{eq:condg_obj} \max ~& \mathbb{E}_{X\sim G_\theta(\cdot|y)}[\sum_{t=1}^{T} \log G_\theta(x_t|X_{1:t\!-\!1}, y) \cdot R_{D_\phi}^{G_\theta}(X_{1:t\!-\!1}, x_t, y)] \\ 
	\label{eq:condd_obj} \max ~& \mathbb{E}_{(X, y)\sim p_{d}}[\log D_\phi(X, y)]\!+\! \mathbb{E}_{X\sim G_\theta(\cdot|y)}[\log (1\!-\! D_\phi(X, y))]
\end{align}
\end{small}
where, 
\begin{small}
\begin{equation}
    \begin{aligned}
	\label{eq:cond_qvalue}
	R_{D_\phi}^{G_\theta}(s\!=\! X_{1:t\!-\!1}, y, a\!=\! x_t)=
	\frac{1}{K} \sum_{k=1}^K D_\phi(X_{1:T}^k, y) \\
	, X_{1:T} \in \text{MC}^{G_\theta}(X_{1:t}, y)
	\end{aligned}
\end{equation}
\end{small}

\begin{algorithm}
\algsetup{linenosize=\small} \scriptsize
	\caption{Medical Text Generative Adversarial Network}
	\begin{algorithmic}[1]
	  \label{algo:main}
	  \REQUIRE{Generative network $G_\theta$; Rollout network $G_\beta$ and update rate $\alpha$; Discriminative network $D\phi$; Real EMR text dataset $S^+ = \{X_{1:T}, y\}$}
	  \STATE Initial $G_\theta, D_\phi$ with random weights $\theta, \phi$
	  \STATE Pre-train $G_\theta$ using MLE on $S^+$
	  \STATE $\beta \gets \theta$
	  \STATE Generate samples $S^-$ using $G_\theta$ for training $D_\phi$
	  \STATE Pre-train $D_\phi$ by \eqref{eq:condd_obj} on ${S^+, S^-}$
	  
	  \FOR{total iterations}
		\FOR{g-steps} 
		  \STATE Assign random disease features $y$, generate sequences $X_{1:T}=(x_1, \dots, x_T) \sim G_\theta(X|y)$ 
		  \FOR{$t$ in $1:T$}
			\STATE Compute rewards $R_{D_\phi}^{G_\beta}$ by \eqref{eq:cond_qvalue}
		  \ENDFOR
		  \STATE Update generator parameters via \eqref{eq:condg_obj}
		\ENDFOR
  
		\FOR{d-steps}
		  \STATE Generate negative examples $S^-$ using current $G_\theta$ and combine with given positive examples $S^+$
		  \STATE Update discriminator parameters via \eqref{eq:condd_obj} for $k$ epochs
		\ENDFOR
		\STATE $\beta \gets (1 - \alpha) \theta + \alpha \beta$
	  \ENDFOR
	\end{algorithmic}
\end{algorithm}

\subsection{Generator Specification}
We use LSTM, an improved variant of RNN, as our generative model. It is worth noticing that other RNN variants, such as GRU, can also be used as the generative model. A typical LSTM generator is shown in Fig. \ref{fig:generator}. RNN maps the input word embedding representation $x_t$ to the hidden state $h_t$ with an update function $f$ recursively, i.e. $h_t = f(h_{t-1}, x_t)$. The improvement of LSTM is that it uses three well-designed gates to implement $f$. 
Specifically, the Forget Gate decides how much information to be abandoned from the cell state; the Input Gate decides what information will be saved from the cell state; the Output Gate filters the input and decides the new hidden state. 
To impose the conditional constraint, the disease features are fed into the model as an additional input at every generation step. 


\subsection{Discriminator Specification} 
The discriminator needs to execute a two-class text classification task. In this paper, we test three types of classifiers in the experiment section: fastText, convolutional neural network (CNN) and bidirectional recurrent neural network (BiRNN) with attention mechanism.
\paragraph{fastText}
FastText is a simple and efficient linear model for text classification\cite{fasttext}. In fastText, each word in a sentence is embedded into a word representation, and these word representations are then averaged to obtain a text representation, which is eventually fed into a linear classifier to output the probability that the sentence comes from the real distribution. In order to take the word order into account, we also use additional n-gram features as the input.

\paragraph{CNN}
Convolutional neural network is the main framework for solving computer vision problems, but it has also shown great performance in text classification recently \cite{textcnn}. Given an input sentence $X = \{x_1, \dots, x_T\}$, we also embed each word to a vector representation $\varepsilon_{1:T} \in \mathbb{R}^{T\times k}$ first, where $k$ is the dimension of word embeddings. Then, we perform convolutional operations on the sentence matrix with different sizes and numbers of filters, of which the second dimension is always $k$. For a filter with size $h$, we can get a feature map $\tilde{c} = \{ c_1, \dots, c_{T-h+1} \}$,  where $c_i = f(\mathbf{w} \otimes \varepsilon_{i:i+h-1} + b)$. Finally we perform max-pooling on $\tilde{c}$ to get $\hat{c} = \max\{\tilde{c}\}$ and fully-connected on $\hat{c}$ to get the final result. Similar to the discriminator in SeqGAN, we also use a residual highway structure before final fully-connected layers to enhance the predictive performance.

\paragraph{BiRNN-attention}
RNN should be the most direct model to conduct text classification task. We can simply use hidden states of the last time step to predict labels. In practice, we use the bidirectional LSTM structure and the attention mechanism to enhance the performance \cite{birnn}. The bidirectional structure uses both pre-word and post-word contexts, rather than only pre-word contexts. The attention mechanism considers that different time steps have different contributions to the target task. Thus, it assigns normalized weights to the hidden states of each time step. The attention mechanism can be formulated as \eqref{eq:bilism}.

\begin{equation}
	\label{eq:bilism}
		\left\{
			\begin{aligned}
				u_t &= \text{tanh}(W_w h_t + b_w) \\
				\alpha_t &= \frac{\exp(u_t^Tu_w)}{\Sigma_t \exp(u_t^Tu_w)} \\
				s &= \sum_t \alpha_t h_t
			\end{aligned}
		\right.
\end{equation}
where $W_w$ and $b_w$ denotes weights and bias of the attention layer, $u_w$ denotes weights of the fully-connected layer to compute $\alpha$. $\alpha_t$ is the weights of outputs at each time step, and $s$ is the final output.

Among all discriminators, the conditional constraint $y$ is fed into the final fully-connected layer as an additional input. 

\subsection{Approaches to Stable Adversarial Training}
In the adversarial text generation, the training usually suffers from two main problems \cite{ntg-review}. One is the gradient vanishing problem, which means if D is much stronger than G, the generated samples will always obtain almost 0 reward, and updates of the generator will nearly stop. The other is the mode collapse problem, which is caused by the REINFORCE algorithm. The generator usually tends to produce short repeated parts to earn high evaluation from the discriminator, which makes the overall quality and diversity of generated samples pretty low. To alleviate these problems, we use several approaches to enhance the stability of adversarial training. 

\paragraph{Rescale Rewards}
To alleviate the gradient vanishing problem, a straight method is to use rescaled scores as reward signals. In the experiment section, we test two rescaled methods. The first one is proposed in MaliGAN \cite{maligan}: $R = \frac{D}{1-D}$, and we call it ODA (Optimal Discriminator Activation) for short. The second one is BRA(Bootstrapped Ranking Activation), which is proposed in LeakGAN \cite{leakgan}: $R = \sigma(\delta \cdot (0.5 - \frac{\text{rank}(i)}{B}))$, where rank($\cdot$) denotes the sequence's high-to-low ranking in the batch, $\delta$ is the activation smoothness hyper-parameter, $B$ is the batch size, and $\sigma(\cdot)$ is a non-linear function. ODA assumes that the discriminator is optimal, while BRA uses the ranking information in each batch to rescale rewards. These two methods do not require any modification on the model structure. Furthermore, we can introduce an action-independent baseline $b$, and use $R - b$ to replace $R$ to guide the update of the generator's parameters. 

\paragraph{Teacher Forcing}
Considering that the generator does not get access to real samples directly, we conduct one step of teacher forcing (MLE) after one step of adversarial training. In fact, the difference between them is that teacher forcing uses texts from real data and the value of rewards is 1, while adversarial training uses texts generated by $G_\theta$ are rewards are given by discriminator. In practice, we find teacher forcing can effectively alleviate the mode collapse problem.

\paragraph{Delayed Rollout Network}
From the RL viewpoint, the generator can be viewed as a \emph{actor} and the discriminator can be viewed as a \emph{critic} \cite{dqn}. To improve the training convergence and stability, we build up a copy of generator $G_\beta$ that is soft-updated with $G_\theta$ to sample sentences in MC search. This delayed rollout network has smaller changes in parameters, which makes the training more stable.

\section{Experiment}
In this section, we first describe the EMR text dataset that we used. Then, we describe the implementation details of our model. To ensure mtGAN achieves the best performance, we conduct experiments with different rescale methods and discriminator structures. To test the effectiveness of our model, we design the micro-level, macro-level and application-level experiments, from which we can see mtGAN has better performance than other baselines.

\subsection{Dataset}
The EMR texts we used are in Chinese, but it is worth noticing that the difference between EMR texts in different languages only lies in the approaches of data pre-processing. After EMR texts are converted into sequences of word embedding vectors, we can simply apply our model to them.

We collected 2216 EMR texts from the respiration department of a hospital to construct the dataset. Original EMR texts include personal information, chief complaint, history of present and past illness and admission diagnosis. To protect the privacy of patients, we remove sensitive information such as person names and place names, and use history of present illness as input sequences and admission diagnosis as sequence tags, or the conditional constraint of generation. In practice, we use two tags: pneumonia and lung cancer. We segment words in each EMR text with the jieba package \cite{jieba} with an additional medical dictionary, and in the end, the dictionary of this dataset includes 7674 words. We cut the first 40 words of each EMR text as the input to the model for convenience, as it will not take a long time to train the model but can also generate informative results. After removing invalid and repeated data, the dataset is split into training, validation and test set with the proportion of 0.7, 0.1, 0.2 separately. A pneumonia example and a lung cancer example of real EMR text data are shown in Tab. \ref{tab:real_example}.

\begin{CJK}{UTF8}{gbsn}
\begin{table}[tb]
	\caption{Examples of original real EMR text}
	\begin{center}
		\begin{tabular}{lp{6cm}}
		\hline
		Type & Examples \\
		\hline
		{Pneumonia}  & 患者 病情 平稳 ， 偶有 咳嗽 ， 咳痰 ， 为 白色 痰 ， 无 明显 喘 憋 ， 无 痰中 带血 ， 无发热 。 今日 为行 进一步 气管镜 下 治疗 入院 。 发病 以来 ， 神清 清楚 ， 精神 可 \\
		\hline
		{Lung Cancer} & 患者 1 年 多 前 出现 咳嗽 ， 咳痰 费力 ， 无 明显 气促 ， 在 外院 考虑 为 ＂支气管炎＂， 给予 口服 头孢类 抗生素 ， 服药 后 疗效 不佳 ， 行 胸部 CT 示 右肺 占位 。 \\
		\hline
		\end{tabular}
		\label{tab:real_example}
	\end{center}
\end{table}
\begin{table}[tb]
	\caption{Examples of generated synthetic EMR text}
	\begin{center}
		\begin{tabular}{lp{6cm}}
		\hline
		Type & Examples \\
		\hline
		{Pneumonia}  & 患者于1周前无明显诱因出现咳嗽，咳白色粘痰，伴活动后加重，休息后可缓解，间断服用镇咳药物等治疗，未行正规诊治。 \\
		\hline
		{Lung Cancer} & 患者10多年前开始出现咳嗽、咳痰，痰中带血，当地医院查胸部CT示纵隔肿大淋巴结，右肺下叶结节，后行气管镜时治疗收入院。 \\
		\hline
		\end{tabular}
		\label{tab:fake_example}
	\end{center}
\end{table}
\end{CJK}


\subsection{Implementation Details}
Our model is implemented in Python with TensorFlow library. For the generator, the word embedding dimension and the hidden state dimension of LSTM cell are both set to 32. The word embedding layer are jointly trained with the generator. For the discriminator, the word embedding dimension is also set to 32. To alleviate overfitting, we add a dropout layer with 0.2 drop rate before the fully-connected layer and L2 regularization with $\lambda = 0.1$. The word embedding layer is fixed with pretrained weights by Word2Vec. We first pretrain the generative model 1000 epochs by MLE and pretrain the discriminative model 500, 100, 100 epochs for fastText, CNN and BiRNN-attention separately by minimizing the cross-entropy between synthetic samples and real samples. In the adversarial training, we update the generator five steps and then update the discriminator five steps. Each REINFORCE step is companied with one step of teacher forcing. The total number of adversarial training epochs is 100. A pneumonia example and a lung cancer example of generated synthetic EMR text data are shown in Tab. \ref{tab:fake_example}. 

\subsection{Micro-Level Evaluation}
We test different discriminator structures and rescale methods with micro-level evaluation metrics in this subsection. To evaluate the model performance, we use the negative log-likelihood on the test set (NLL-test) and self-BLEU \cite{benchmark} as evaluation metrics. NLL-test evaluates the model's capacity to fit the real data and self-BLEU is a metric to evaluate the diversity of generated synthetic sentences. The calculation of NLL-test can be formulated as:
\begin{equation}
	\text{NLL}_{test} = -\mathbb E_{\mathbf{x}\in S_{\text{test}}}[\sum_{t=1}^{T} \log G_\theta(x_t|X_{1:t-1})]
\end{equation}
For the self-BLEU metric, since BLEU aims to evaluate how similar two sentences are, we can calculate BLEU between each sentence and the rest in a group of generated sentences, which indicates how this sentence resembles with others. The average BLEU score of each generated sample is defined as self-BLEU, which evaluate the diversity of current generative model. The higher self-BLEU score indicates the less diversity of generated samples. The calculation of self-BLEU can be formulated as:
\begin{equation}
	\text{self-BLEU} = \mathbb E_{\mathbf{x}\in S_{\text{test}}}[\text{BLEU}(X|\complement_{S_\text{test}}(X))]
\end{equation}

\begin{figure}[tb]
    \centering
    \includegraphics[width=8cm, height=5cm]{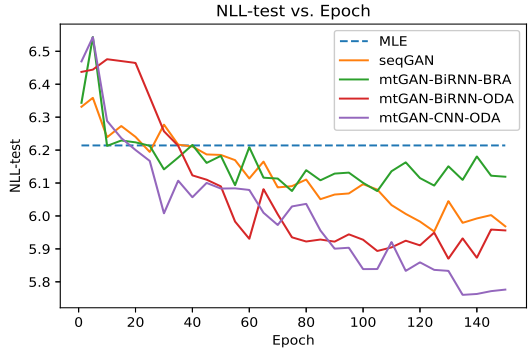}
    \caption{NLL-test results with different discriminators and rescale reward methods}
    \label{fig:nll-test}
\end{figure}

\begin{table}[tb]
	\caption{Micro-Level Experiment Results}
	\begin{center}
		\begin{tabular}{c|c|c}
		\hline
		Model & NLL-test & self-BLEU \\
		\hline
		MLE & 6.2141 & 0.9270 \\
		SeqGAN & 5.9685	 & 0.9267  \\
		mtGAN-BiRNN-BRA & 6.1191 & 0.9155  \\
		mtGAN-BiRNN-ODA & 5.9561 & 0.9209 \\
		mtGAN-CNN-ODA & \textbf{5.7764} & \textbf{0.9182} \\
		\hline
		\end{tabular}
		\label{tab:micro_eval}
	\end{center}
\end{table}

The NLL-test scores of MLE training and adversarial training with different discriminative models and rescale reward methods are shown in Fig. \ref{fig:nll-test}. SeqGAN is an extended conditional model based on the original version and does not use any rescale reward methods, which we also view as a baseline model together with MLE. It can be seen that the rescale reward method ODA is better than BRA and no rescale method. Comparing the curves of MLE, SeqGAN, mtGAN-BiRNN-ODA and mtGAN-CNN-ODA, it is obvious that our mtGAN with CNN discriminator can achieve a better NLL-test score than BiRNN discriminator. In addition, all models trained by adversarial training are better than the model trained by MLE. The fastText discriminator will assign improper reward signals, which lead to the failure of training generator. 

The NLL-test and self-BLEU scores are shown in Tab. \ref{tab:micro_eval}. The mtGAN-CNN model can achieve highest NLL-test score and relatively low self-BLEU score, which indicates that our model has the strong ability to fit the real data and generate diverse examples at the same time. We will use mtGAN-CNN-ODA as our main model to conduct the following evaluation.

\subsection{Macro-Level Evaluation}
For the macro-level evaluation, we conduct an adversarial evaluation experiment similar to \cite{b11} to fairly evaluate the whole quality of synthetic EMR texts. The idea of adversarial evaluation resembles the idea of Turing test. In the adversarial evaluation, we train a separate machine evaluator in place of the human evaluator to distinguish real EMR texts and generated synthetic EMR texts. We report Adversarial Success (AdverSuc) on the test set, which is the fraction of instances that generated samples can fool the evaluator. Higher scores of AdverSuc indicates that the generated synthetic samples are more similar to real samples. However, the adversarial evaluation is model-dependent, a poor discriminative model can also lead to a low accuracy of the evaluator. Thus, we also set up three manually-designed situations to measure the capacity of the evaluator. 
\begin{enumerate}
    \item Randomly split real EMR texts as positive examples and negative examples. An ideal evaluator should give an accuracy of 0.5.
    \item Randomly split generated EMR texts as positive examples and negative examples. An ideal evaluator should also give an accuracy of 0.5.
	\item Use real EMR texts as positive examples and random generated EMR texts as negative examples. An ideal evaluator should give an accuracy of 1.0.
\end{enumerate}
We report the absolute difference value between the accuracy of the evaluator and the ideal accuracy in such three experiments as evaluator reliability error (ERE). The lower ERE value indicates the higher model reliability. We train a separate discriminator in the adversarial evaluation experiment. The AdverSuc and ERE scores of MLE, SeqGAN and our mtGAN are shown in Tab. \ref{tab:adver_eval}. Our mtGAN model can achieve the highest AdverSuc and lower meanERE than MLE at the same time, which indicates the synthetic EMR texts generated by our model have the higher quality than samples generated by MLE, and this result is reliable. It is hard to compare the results between SeqGAN and mtGAN, as we are not sure what mtGAN's AdverSuc will be if the evaluator is more reliable. However, the AdverSuc of SeqGAN is pretty low, which implies the quality of synthetic EMR texts is not satisfactory.

\begin{table}[tb]
	\caption{Macro-Level Experiment Results}
	\begin{center}
		\begin{tabular}{c|c|c|c|c|c}
		\hline
		Model  &AdverSuc	&ERE1	&ERE2	&ERE3	&meanERE \\
		\hline
		MLE	   	&0.3007		&0.0068	&0.0676	&0.3446	&0.1396 \\
		SeqGAN	&0.1351		&0.0203	&0.0270	&0.1081	&0.0518 \\
		mtGAN	&\textbf{0.3041}	&0.0811	&0.0270	&0.2804	&0.1295 \\
		\hline	
		\end{tabular}
		\label{tab:adver_eval}
	\end{center}
\end{table}

\subsection{Application-Level Evaluation}
For the application-level evaluation, we design a practical classification experiment to test if the samples generated by our model can be used as the source of data augmentation and help other machine learning tasks. The labels are pneumonia and lung cancer, which is set same to the conditional constraint used in the model training. It is worth noticing that during the whole training process (from pretraining to adversarial training), our model does not get access to the test set, so we can assign labels to our model to generate corresponding samples, and use these synthetic EMR texts to train a classifier and evaluate it on the test set. Ideally, using generated samples to train a classifier should achieve similar performance to using real samples, and adding generated samples to real samples should achieve a higher score as the result of data augmentation. 
\begin{table}[tb]
	\caption{Application-Level Experiment Results}
	\begin{center}
		\begin{tabular}{c|c|c|c}
		\hline
		\diagbox{Model}{Accuracy}{Data source} & Real & Synthetic & Mix \\
		\hline
		MLE & 0.7500 & 0.7432 & 0.7568 \\
		SeqGAN & 0.7500	 & 0.6959 & 0.7095  \\
		mtGAN & 0.7500 & 0.7432 & \textbf{0.7635} \\
		\hline
		\end{tabular}
		\label{tab:disease_cls}
	\end{center}
\end{table}
The application-level evaluation results are shown in Tab. \ref{tab:disease_cls}. It can be seen that the classifier can achieve the best accuracy 76.35\% with the mix of real EMR text and synthetic EMR text. Thus, the synthetic EMR texts generated by our model do have similar properties to real EMR texts, and can be used as a data augmentation approach to assist other tasks such disease classification task. 

\section{Conclusion}
In this paper, we propose a conditional model mtGAN to generate synthetic EMR texts. This method solves the privacy problem and the insufficient and imbalance samples problem naturally. The micro-level, macro-level and application-level evaluation demonstrate that our model can generate more realistic EMR texts, which has a wide range of applications. The future work will focus on the hidden representation of EMR texts, which will help us impose more direct control to the generation process and also help us analyze the similarity of different EMR texts to better understand diseases.

\section*{Acknowledgment}
This work is supported in part by National Key R\&D Program of China (2008YFC0910401, 2018YFC0910404) and NSFC grants 61721003, 11801301.

\end{document}